\documentclass{article}
\usepackage{amssymb}
\usepackage{amsmath}

\usepackage[preprint]{corl_2025} 

\usepackage{booktabs}
\usepackage{algorithm}
\usepackage{algpseudocode}
\usepackage{multirow}
\usepackage{graphicx}
\usepackage{wrapfig}
\usepackage{array}
\usepackage{float}
\usepackage{bbm}
\usepackage{tablefootnote}
\usepackage{subcaption}
\usepackage{comment}
\newcommand{\drule}{\specialrule{0.2pt}{1pt}{1pt}%
            \specialrule{0.2pt}{0pt}{\belowrulesep}%
            }
  {\begin{list}{}%
          {\setlength{\leftmargin}{#1}}%
          \item[]%
  }
  {\end{list}}

\usepackage{pifont}
\usepackage{xcolor}
\usepackage{pdfrender}
\usepackage{colortbl}
  
\usepackage{kotex}

\definecolor{applegreen}{rgb}{0.55, 0.71, 0.0}

\definecolor{forestgreen}{rgb}{0.13, 0.55, 0.13}

\title{Uncertainty-aware Accurate Elevation Modeling for Off-road Navigation via Neural Processes\vspace*{-0.4cm}}

\author{
Sanghun Jung\\
University of Washington\\
\texttt{shjung13@cs.washington.edu}
\And
Daehoon Gwak\\
KAIST\\
\texttt{daehoon.gwak@kaist.ac.kr}
\And
Byron Boots\\
University of Washington\\
\texttt{bboots@cs.washington.edu}
\And
James Hays\\
Georgia Institute of Technology\\
\texttt{hays@gatech.edu}
}

\begin{document}
\maketitle


\vspace*{-0.3cm}
\vspace{-0.7cm}
\begin{abstract}
Terrain elevation modeling for off-road navigation aims to accurately estimate changes in terrain geometry in real-time and quantify the corresponding uncertainties.
Having precise estimations and uncertainties plays a crucial role in planning and control algorithms to explore safe and reliable maneuver strategies.
However, existing approaches, such as Gaussian Processes (GPs) and neural network-based methods, often fail to meet these needs.
They are either unable to perform in real-time due to high computational demands, underestimating sharp geometry changes, or harming elevation accuracy when learned with uncertainties.
Recently, Neural Processes (NPs) have emerged as a promising approach that integrates the Bayesian uncertainty estimation of GPs with the efficiency and flexibility of neural networks.
Inspired by NPs, we propose an effective NP-based method that precisely estimates sharp elevation changes and quantifies the corresponding predictive uncertainty without losing elevation accuracy.
Our method leverages semantic features from LiDAR and camera sensors to improve interpolation and extrapolation accuracy in unobserved regions. 
Also, we introduce a local ball-query attention mechanism to effectively reduce the computational complexity of global attention by 17\% while preserving crucial local and spatial information.
We evaluate our method on off-road datasets having interesting geometric features, collected from trails, deserts, and hills.
Our results demonstrate superior performance over baselines and showcase the potential of neural processes for effective and expressive terrain modeling in complex off-road environments.
\end{abstract}
\vspace{-0.3cm}

\vspace{-0.1cm}
\vspace{-0.1cm}
\vspace{-0.4cm}
\section{Introduction}
\begin{figure}[t!]
  \centering
  \includegraphics[width=0.8\textwidth]{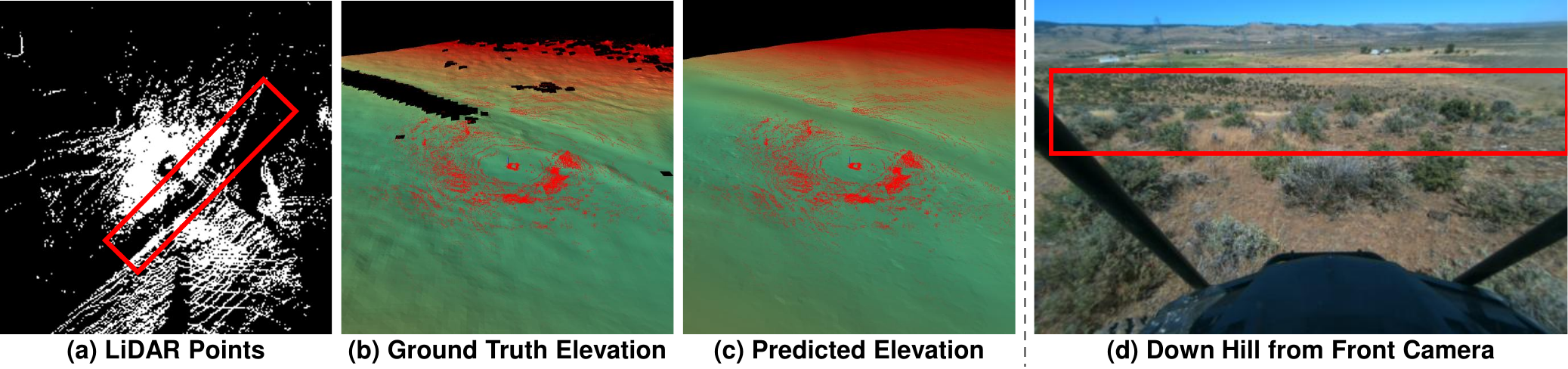}
  \vspace{-0.2cm}
  \caption{(Left group) An example of LiDAR points on a ditch and its ground truth and predicted elevations from our model. (Right) An example image from a forward-facing camera looking downhill. The terrain looks continuous, but there is a significant downhill slope in front of the vehicle.}
  \vspace{-0.9cm}
\label{fig:neg_obstacle_example}
\end{figure}

\vspace{-0.3cm}
Accurate terrain elevation modeling is a long-standing problem in robotics~\cite{elevation_vhr, pixel_to_elev, elevation1, elevation2_tal, elevation3, elevation4, elevation5}. Reliable ground estimations play a crucial role in determining a vehicle's maneuvering strategies~\cite{planning_under_uncert1, planning_under_uncert2, planning_under_uncert3, lrn} for off-road navigation. For example, negative obstacles (e.g., ditches or craters) are best avoided, but when they must be traversed, the vehicle needs accurate elevation estimates to align itself and plan a trajectory that negotiates the terrain safely. However, such negative obstacles are often not visible in LiDAR and camera sensors, especially when observed from a distance due to limited viewing angles. Instead, they appear as a `narrow gap' between LiDAR rays in point clouds or as `continuous ground' in images, as illustrated in Fig.~\ref{fig:neg_obstacle_example}. This invisible nature complicates their prediction and frequently leads to underestimation, such as misinterpreting ditches as shallower depressions or cliffs as downhill slopes.
We aim to accurately estimate the predictive distribution of ground elevation in each cell within bird's-eye-view (BEV) grids while addressing challenging cases such as negative obstacles, which require extrapolation beyond direct observations.

From LiDAR observations at each timestep, ground elevations can be estimated by taking the minimum height value in each BEV cell. However, given the sparse nature of LiDAR sensors and their limited viewing angles, there is a crucial need to effectively interpolate and extrapolate these observed `context' points to unobserved `target' points. Early research predominantly utilized Gaussian Processes (GPs)~\cite{gp, elevation_gp, elevation_gp2, elevation_gp3, elevation_gp4, elevation_gp5, elevation_gp6, elevation_gp7} and Bayesian Generalized Kernels (BGKs)~\cite{bgk, elevation_bgk, elevation_bgk2, elevation_bgk3} to address this task. They provide a solid mathematical foundation for making predictions and understanding their associated uncertainties. Nevertheless, their flexibility is restricted due to predefined kernel functions and fixed prior distributions. Moreover, their computational demands often prevent the use of real-time applications, which are crucial for agile, high-speed off-road autonomy.

More recently, deep neural network-based methods~\cite{bevnet, terrainnet, pixel_to_elev, elevation3, elevation2_tal, vstrong} have emerged, exhibiting strong learning capabilities from extensive datasets and thus better generalization across varied terrain characteristics. Although some methods~\cite{bevnet, terrainnet} produce promising predictions on relatively flat terrains, they often underestimate sharp geometric features and inadequately model negative obstacles. This shortcoming leads to overly optimistic planning and control strategies, which can potentially cause vehicle rollovers or falls from cliffs. 
Additionally, accurate uncertainty estimations are essential for identifying and avoiding regions where predictions are less confident.
However, we observe that training these models to estimate uncertainty often accompanies degradation in elevation accuracy.

To overcome the limitations, we propose leveraging Neural Processes (NPs)~\cite{np, anp, bnp, convcnp}, which learn adaptive, data-driven representations without relying on predefined kernels or fixed priors, thereby enhancing flexibility in modeling diverse terrain features. NPs inherently provide accurate uncertainty estimation without compromising prediction accuracy, and they exhibit strong capability for capturing sharp changes in elevation. Also, NPs scale effectively for real-time deployment due to their computational efficiency compared to GPs.
In this paper, we aim to address the following major challenges:  1) precisely estimating sharp elevation changes, 2) improving interpolation and extrapolation accuracy in unobserved regions, and 3) quantifying associated uncertainty.
While the associated uncertainties are inherently learned through NP frameworks, the remaining challenges require special attention.

\noindent\textbf{Improving Inter-/Extrapolation.}
Analogous to existing literature~\cite{np, anp, elevation_gp, elevation_gp2}, `context points' (i.e., estimated ground height from LiDAR) can be the sole basis for elevation prediction at any other unobserved target point. 
However, they can be sparse and lack semantic information. Therefore, context-point-only methods struggle with extrapolating context points, often failing significantly in unobserved regions.
Thus, we incorporate additional semantically informative features from both the camera and LiDAR sensors and condition the context and target points on them.
This \emph{semantic-conditioned NP} enables the prediction of reasonable elevations beyond the observed areas.

\noindent\textbf{Estimation of Sharp Elevation Changes.}
An attention mechanism~\cite{attention} is an effective solution to preserve sharpness in predictions~\cite{anp} since it promotes models to learn self-correlations among observations.
However, our observations cover regions larger than 100 m × 100 m. Thus, using global attention requires a significant memory, which is often unfeasible for onboard computing in robots.
To address this, we propose a ball-query attention mechanism where only observations within a local $\epsilon$-ball are considered for attention operations.
Applying it reduces the number of floating-point operations (FLOPS) and inference time by 17\% and 36\%, while preserving crucial local and spatial information.

To preserve temporal consistency, we aggregate LiDAR points and update image features using non-parametric Bayesian updates without requiring additional training. We validate our method on off-road datasets collected from grassy hills in California (i.e., CA Hills), the Mojave Desert with on- and off-trail runs (i.e., Mojave Desert), and dirt and grassy terrain with diverse negative obstacles in Ellensburg, Washington (i.e., Ellensburg). Our method outperforms existing approaches in terms of elevation, slope, and curvature accuracy, while also providing meaningful uncertainty estimates.

To summarize, our contributions are:
\vspace{-0.3cm}
\begin{itemize}
    \item We present precise elevation estimates with associated uncertainties by introducing a novel semantic-conditioned Neural Processes with temporal aggregation.
    \vspace{-0.1cm}
    \item We introduce a ball-query attention mechanism that effectively reduces computational demands while maintaining necessary local, spatial information.
    \vspace{-0.1cm}
    \item We demonstrate the effectiveness of our method on varied terrain characteristics, collected from trails, deserts, and hills.
\end{itemize}
\vspace{-0.3cm}

\vspace{-0.2cm}
\section{Method}
\vspace{-0.2cm}
This section first explains the problem definition and ANPs~\cite{anp} in the preliminary section, then describes how we obtain robust and temporally-aggregated semantic features from LiDAR and cameras, and finally presents semantic-conditioned NPs with ball-query attention.
\vspace{-0.1cm}
\subsection{Preliminary}
\vspace{-0.1cm}
\noindent\textbf{Problem Formulation.}
The terrain elevation modeling task aims to predict the ground elevation $h_{x, y}$ at location $ (x, y) $ in BEV grids.
Given ego-centric LiDAR readings, $\{x_i, y_i, z_i\}_N$, we first crop the readings with predefined ranges: $x$: [-51.2 m, 51.2 m] and $y$: [-51.2 m, 51.2 m].
The $(x, y)$ values of cropped LiDAR readings will be binned into each grid with a resolution of $R=0.4$, resulting in 256$\times$256 BEV grids.
For each center point of grid, we estimate the ground elevation at BEV cell $(x, y)$ by $\hat{h}_{x,y} := \min_{\{x^\prime, y^\prime\}} z,$ where $\forall {(x^\prime, y^\prime)} \in [x - \frac{R}{2}, x + \frac{R}{2}] \times [y - \frac{R}{2}, y + \frac{R}{2}]$, if there are any LiDAR hits in the cell.
Our task is to predict ground truth $h_{x, y}$s from given estimations $\hat{h}_{x, y}$s, LiDAR readings, and camera inputs.

\begin{figure}[t!]
  \centering
  \includegraphics[width=0.80\textwidth]{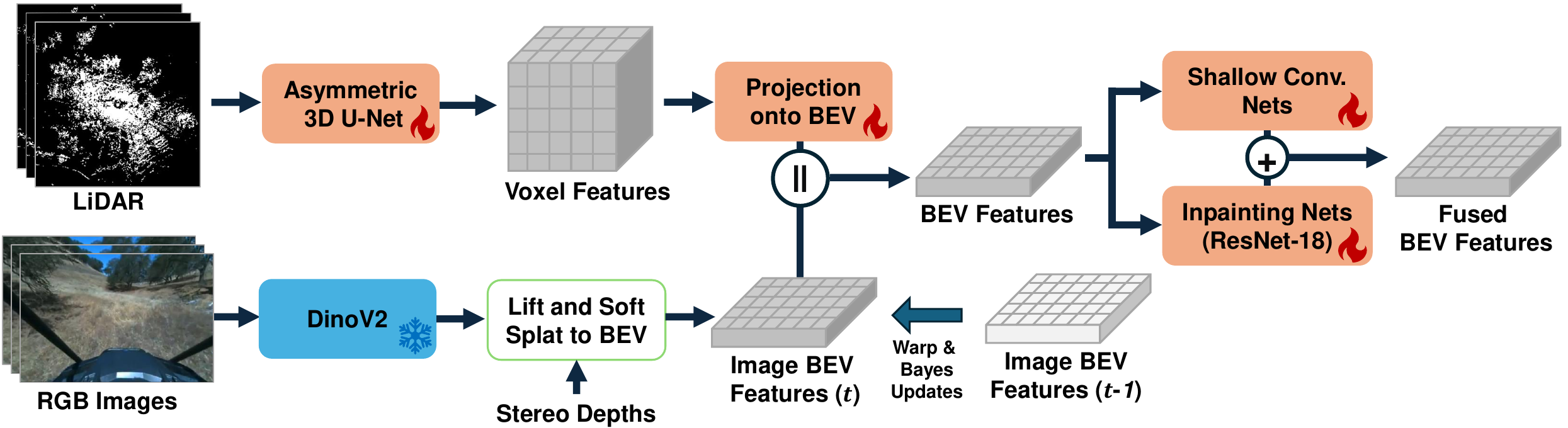}
  \vspace{-0.30cm}
  \caption{Overview of semantic extraction from LiDAR and camera sensors. $||$ operation denotes the concatenation of two features. We use the Asymmetric 3D U-Net~\cite{asymm} and DinoV2~\cite{dinov2} for LiDAR and image features. We lift the image features to 3D space using stereo depth maps and project them onto BEV space. Image BEV features are then temporally aggregated using Bayesian updates. Concatenated LiDAR and Image features are passed to shallow convolutional neural networks and inpainting networks to obtain fused features.}
  \vspace{-0.5cm}
\label{fig:semantic_features}
\end{figure}
\noindent\textbf{Attentive Neural Processes.}
In this work, we adopt ANPs~\cite{anp} as our base framework. Formally, given a set of context points $(\mathbf{X}_C, \hat{\mathbf{H}}_C) := \{(x_i, y_i), \hat{h}_i\}_{i \in C}$ from LiDAR readings and target points $\mathbf{X}_T := \{(x_j, y_j)\}_{j \in T}$ sampled from BEV grids, ANPs learn to predict the conditional distribution $p(\mathbf{H}_T | \mathbf{X}_T, \mathbf{X}_C, \hat{\mathbf{H}}_C)$. ANPs model the conditional distribution through a global latent variable $\mathbf{z}$ sampled from aggregated context representation $\mathbf{s}_C$ and a deterministic context representation $\mathbf{r}_C$. Specifically, the conditional predictive distribution is given by:
\begin{equation}
p(\mathbf{H}_T | \mathbf{X}_T, \mathbf{X}_C, \hat{\mathbf{H}}_C) = \int p(\mathbf{H}_T | \mathbf{X}_T, \mathbf{r}_C, \mathbf{z}) q(\mathbf{z} | \mathbf{s}_C) d\mathbf{z},
\end{equation}
where $q(\mathbf{z} | \mathbf{s}_C)$ represents the variational prior inferred from the context. 
The permutation-invariant encoder $\mathbf{s}_* := \mathbf{s}(\mathbf{X}_*, \mathbf{H}_*)$ aggregates context/target points and predicts factorized Gaussian distribution $\mathbf{s}_*$, where the latent representation $\mathbf{z}$ is sampled from.
The latent representation encapsulates global spatial correlations and uncertainty. A deterministic representation $\mathbf{r}_C$ is obtained from a permutation-invariant function $\mathbf{r}(\mathbf{X}_C, \hat{\mathbf{H}}_C, \mathbf{X}_T)$ using global attention. The decoder then uses $\mathbf{z}$ and $\mathbf{r}_C$ to generate predictions for the target points from $\mathbf{X}_T$.
The encoders and a decoder are trained to maximize the Evidence Lower Bound (ELBO)
\begin{equation}
    \log{p(\mathbf{H}_T | \mathbf{X}_T, \mathbf{X}_C, \hat{\mathbf{H}}_C)} \geq \mathbb{E}_{q(\mathbf{z}|\mathbf{s}_T)}[\log{p(\mathbf{H}_T | \mathbf{X}_T, \mathbf{r}_C, \mathbf{z})}] - D_\text{KL}(q(\mathbf{z}|\mathbf{s}_T)||q(\mathbf{z}|\mathbf{s}_C)),
\end{equation}
with randomly sampled sets of context $C$ and target $T$ using reparameterization tricks~\cite{vae}.

\begin{figure}[t!]
\centering
  \includegraphics[width=0.7\textwidth]{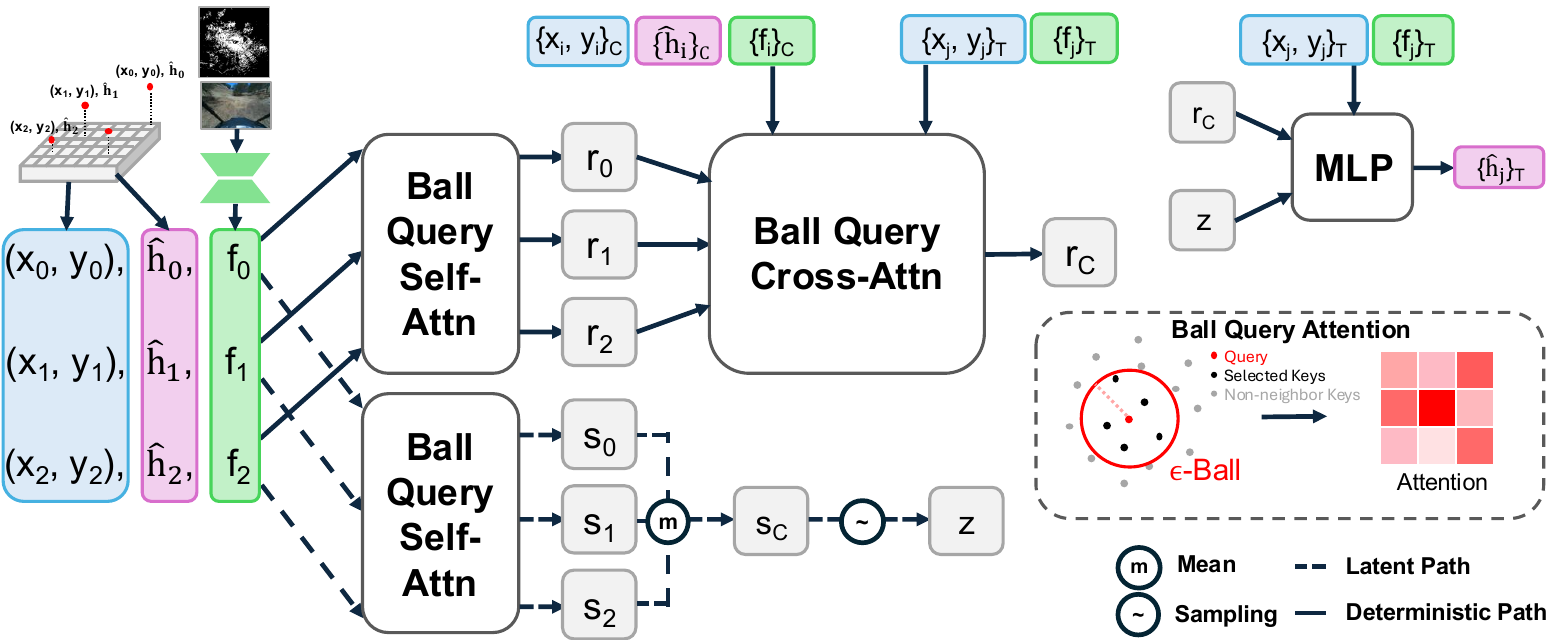}
  \vspace{-0.3cm}
  \caption{Overview of our semantic-conditioned NPs with ball query attention mechanisms. With $(x, y)$ coordinates in BEV space, estimated ground elevation $\hat{h}$, and semantic features $\mathbf{f}$, we predict context representation $\mathbf{r}_C$ and global latent variable $\mathbf{z}$ using ball-query attention mechanisms between context points and across context and target points. Predicted variables are passed to the decoder with target points to predict Gaussian-factorized ground elevations.}
  \vspace{-0.7cm}
\label{fig:np}
\end{figure}

\vspace{-0.4cm}
\subsection{Image and LiDAR Semantic Feature Extraction and Fusion}
\vspace{-0.3cm}
\noindent\textbf{Feature Extraction and Fusion.} Fig.~\ref{fig:semantic_features} illustrates how we extract semantic features from LiDAR and images and fuse them in the BEV space.
We use Asymmetrical 3D U-Net~\cite{asymm} for LiDAR feature extraction and DinoV2~\cite{dinov2} for image feature extraction.
We project the features from each modality onto the BEV space by aggregating voxel features along the z-axis or lifting image features using camera matrices with stereo depths and then soft-splatting them onto the BEV space~\cite{lss, terrainnet}.
Such projected features are then concatenated and passed to the inpainting network to get the fused BEV features.
The inpainting network consists of a \emph{deep} Convolutional Neural Network (CNN) with large receptive fields, which propagates observed features to unobserved regions. This ensures that we generate reasonable elevation predictions in unobserved areas.
However, we observe that such inpainting results in overly smoothed elevation predictions, which lose distinctive features in `observed' areas.
To preserve distinctive features, we introduce an additional path using a \emph{shallow} CNN that fuses features with a limited receptive field, thereby maintaining local details.
The fused results are then added to the inpainted features, which will later be used for NPs.

\noindent\textbf{Temporal Aggregation of Image BEV Features.}
Aggregating observations over time is essential for consistent and precise predictions.
From previous literature~\cite{bevnet, terrainnet}, a recurrent neural network (RNN)-based aggregation has been proposed.
However, we observe several limitations to this approach.
First, it requires a batch of sequences for training, which consumes a significant amount of memory.
Moreover, representing $\hat{h}_{x,y}$ values of observations in the features is essential for precise elevation estimations.
However, the vehicle's elevation $z_\text{ego}$ changes over time, and aggregating information in the feature space can degrade prediction accuracy if we do not compensate for these changes in vehicle elevation correctly.
To minimize this impact, we aggregate LiDAR points instead of features to easily compensate for $z_\text{ego}$, and apply temporal aggregation to image features only, as they contain more semantic and object information rather than direct geometry information.
For image BEV feature aggregation, we use a non-parametric method based on their confidence, which does not require additional training.
During lifting and projection, we find the 3D coordinates for each image pixel, interpolate the lifted features into voxels, and project them down to the BEV space.
We count the number of lifted image pixels for each voxel and project the normalized density $\sigma\in[0,1]$ onto the BEV space.
Using normalized densities as proxies for the probabilities of being correct, we conduct Bayesian belief updates as follows. We use image BEV features from current and previous history, $\hat{\mathbf{f}}_t,\mathbf{f}_{t-1}\in\mathbb{R}^{D\times H \times W}$, and their corresponding probability of being correct, $ \hat{\mathbf{p}}_t, \mathbf{p}_{t-1}\in [0, 1]^{H \times W}$, in the probability simplex.
With a uniform prior assumption and a conditional independence assumption~\cite{bayesian_occupancy}, we obtain a temporally-updated semantic features $\mathbf{f}_t$ by
\begin{gather}
   \mathbf{f}_t = \frac{\hat{\mathbf{p}}_t\,\hat{\mathbf{f}}_t + \mathbf{p}_{t-1} \mathbf{f}_{t-1}}{\hat{\mathbf{p}}_t + \mathbf{p}_{t-1}}, \ \ \ \   
   \mathbf{p}_t = \frac{\hat{\mathbf{p}}_t\mathbf{p}_{t-1}}{\hat{\mathbf{p}}_t\mathbf{p}_{t-1} + (1 - \hat{\mathbf{p}}_t)(1 - \mathbf{p}_{t-1})}.  
\end{gather}
For detailed derivations, please refer to the supplementary materials.

\vspace{-0.3cm}
\subsection{Semantic-conditioned Neural Processes for Elevation Modeling}
\vspace{-0.2cm}
\noindent\textbf{Semantic-conditioned Neural Processes.}
With semantic features, a set of context points is defined as $(\mathbf{X}_C, \hat{\mathbf{H}}_C, \mathbf{F}_C) := \{(x_i, y_i), \hat{h}_i, \mathbf{f}_i\}_{i\in C}$. Accordingly, target points are defined as $(\mathbf{X}_T, \mathbf{F}_T) := \{(x_j, y_j), \mathbf{f}_j\}_{j\in T}$. NPs learn to predict the conditional distribution $p(\mathbf{H}_T|\mathbf{X}_T, \mathbf{F}_T, \mathbf{X}_C, \hat{\mathbf{H}}_C, \mathbf{F}_C)$ by using a global latent variable $\mathbf{z}$ and context representation $\mathbf{r}_C$:
\begin{equation}
p(\mathbf{H}_T | \mathbf{X}_T, \mathbf{F}_T, \mathbf{X}_C, \hat{\mathbf{H}}_C, \mathbf{F}_C) = \int p(\mathbf{H}_T | \mathbf{X}_T, \mathbf{F}_T, \mathbf{r}_C, \mathbf{z}) q(\mathbf{z} | \mathbf{s}_C) d\mathbf{z},
\end{equation}
where $\mathbf{s}_C = \mathbf{s}(\mathbf{X}_C, \hat{\mathbf{H}}_C, \mathbf{F}_C)$, and $\mathbf{r}_C = \mathbf{r}(\mathbf{X}_C, \hat{\mathbf{H}}_C, \mathbf{F}_C, \mathbf{X}_T, \mathbf{F}_T)$.
This strategy enables the model to infer target values not only based on the observed elevation estimates but also to utilize rich semantics obtained from images and the entire point cloud.
Semantic information provides a better understanding of ground geometry and objects, and it improves the accuracy of interpolation and extrapolation where we have semantics but no ground elevation estimates.
Our updated ELBO is:
\begin{equation}
    \log{p(\mathbf{H}_T | \mathbf{X}_T, \mathbf{F}_T, \mathbf{X}_C, \hat{\mathbf{H}}_C, \mathbf{F}_C)} \geq \mathbb{E}_{q(\mathbf{z}|\mathbf{s}_T)}[\log{p(\mathbf{H}_T | \mathbf{X}_T, \mathbf{F}_T, \mathbf{r}_C, \mathbf{z})}] - D_\text{KL}(q(\mathbf{z}|\mathbf{s}_T)||q(\mathbf{z}|\mathbf{s}_C)). \nonumber
\end{equation}

\noindent\textbf{Ball Query Attention.}
As illustrated in Fig.~\ref{fig:np}, our approach uses context points generated from observations that cover a 100 m × 100 m range.
With our default resolution of $R=0.4$ m, we can have a maximum of 256 × 256 points for each context and target.
While existing methods~\cite{anp, tnp} model every relation between them using global attention mechanisms, they are often not applicable to our case due to their high memory requirements.
Instead, we propose using ball-query attention, which significantly reduces memory and computations while preserving spatial relations within local regions.
Specifically, with given query, key, and values $\mathbf{Q} \in \mathbb{R}^{M\times D}, \mathbf{K}, \mathbf{V} \in \mathbb{R}^{N\times D}$, we retrieve a subset of keys and values $\{\mathbf{k}_j, \mathbf{v}_j\}_{j\in\mathcal{B}(\mathbf{q}_i)}$ for each query $\mathbf{q}_i$ in an $\epsilon$-ball $\mathcal{B}(\mathbf{q}_i)$ in the BEV space.
For keys $\mathbf{K}_{\mathcal{B}(\mathbf{q}_i)}$ and values $\mathbf{V}_{\mathcal{B}(\mathbf{q}_i)}$ for each $\epsilon$-ball $\mathcal{B}(\mathbf{q}_i)$, we define scaled dot-product attention for each query $\mathbf{q}_i$ as follows:
\begin{equation}
    \text{\textbf{Attention}}(\mathbf{q}_i, \mathbf{K}_{\mathcal{B}(\mathbf{q}_i)}, \mathbf{V}_{\mathcal{B}(\mathbf{q}_i)}):= \text{Softmax}(\mathbf{q}_i^\intercal\mathbf{K}_{\mathcal{B}(\mathbf{q}_i)}/\sqrt{D})\mathbf{V}_{\mathcal{B}(\mathbf{q}_i)} \in \mathbb{R}^D.
\end{equation}
In practice, we extend this equation to all queries and perform a batch operation for computational efficiency.
We apply a multihead attention mechanism~\cite{attention} based on this dot-product attention. This ball query attention is used for all three attention layers in our architecture.

\vspace{-0.4cm}
\section{Experiments}
\begin{figure}[t!]
\centering
\includegraphics[width=0.8\textwidth]{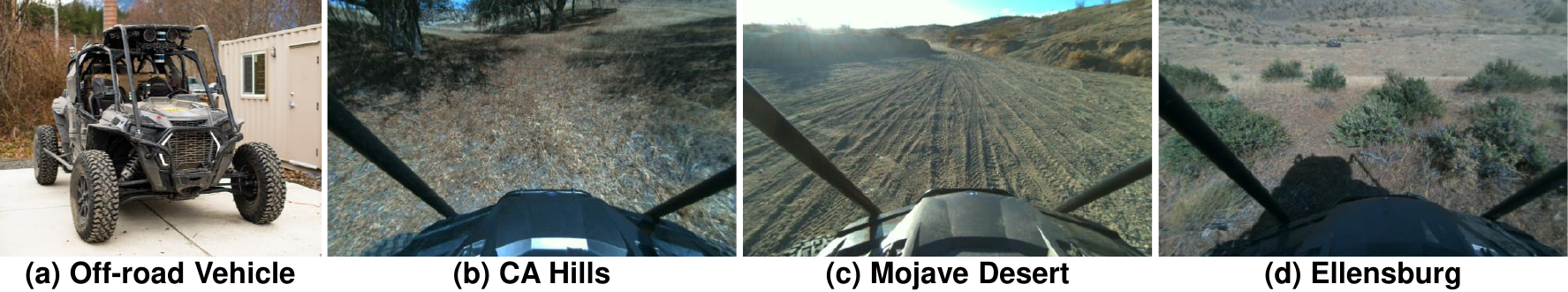}
  \vspace{-0.2cm}
  \caption{(a): Off-road vehicle used for dataset collection.  (b - d): Example images from each site.}
  \vspace{-0.7cm}
\label{fig:dataset}
\end{figure}

\vspace{-0.3cm}
This section describes the experimental settings and evaluation results for three different off-road dataset sequences. The datasets are collected using a Polaris RZR vehicle~\cite{polaris} (Fig.~\ref{fig:dataset} (a)), which can travel at a speed of up to 20m/s (45mph) on off-road terrains. It has three Velodyne 32-beam LiDAR sensors, two at the front and one at the back, and is also equipped with four MultiSense stereo cameras, mounted at the front, left, right, and back. Implementation details are included in the supplementary materials.

\vspace{-0.2cm}
\subsection{Datasets}
\vspace{-0.2cm}
Datasets (Fig.~\ref{fig:dataset} (b - d)) are collected from CA Hills that present grassy up and down hills with tall trees, Mojave desert that has on- and off-trail regions with Joshua trees, small bushes, and rocks, and Ellensburg that has large and small bushes with interesting negative obstacles such as ditches, steep up/down hills, and cliffs.
CA Hills and Mojave sequences are adopted from TerrainNet~\cite{terrainnet}, and the Ellensburg sequence is processed by the same protocol.
Sequences are pose-corrected using an offline SLAM algorithm~\cite{cartographer}, and sensors are synchronized to minimize their time gap.
Elevation ground truths are generated by aggregating 300 LiDAR scans (running at 10 Hz) with a time range of ($t-150$, $t+149$) and calculating the minimum height in each BEV grid cell~\cite{terrainnet}, allowing us to have ground truth elevations even in unobserved regions at time $t$.

\vspace{-0.2cm}
\subsection{Experimental Setup}
\vspace{-0.1cm}
\noindent\textbf{Evaluation Metrics.} We measure mean absolute error (MAE) of 1) elevation predictions (meters), 2) their slopes (\%) along with $x$ and $y$ axes, and 3) their curvature (1/meters).
Slopes are measured by $100\times\frac{\Delta h}{\Delta x}$ and $100\times\frac{\Delta h}{\Delta{y}}$.
Note that the curvatures are approximated using Laplacian filtering.
We measure these metrics separately for observed and unobserved regions.
Observed regions denote BEV cells with current LiDAR observations at each timestep $t$; otherwise, they are unobserved regions.
We expect different levels of precision for observed and unobserved regions.
The models should predict precise elevations in observed areas, while they may only predict smoother, approximate elevations in unobserved regions.

\noindent\textbf{Baselines.}
We compare our method with Deep Gaussian Processes (DeepGPs)~\cite{deepgp}, Hilbert Maps (HMs) with random Fourier features (RFFs)~\cite{hilbert_maps}, Sparse Gaussian Processes (SGP)~\cite{sgp1, sgp2, sgp3}, BEVNet~\cite{bevnet}, TerrainNet~\cite{terrainnet}, and a fused approach of BEVNet and TerrainNet (i.e., Fusion).
SGP takes a representative subset of context points as inducing points to approximate the full context points, reducing the computational complexity of GPs by applying Bayes' rule.
Considering that our LiDAR context points for each frame can cover a 256 $\times$ 256 BEV grid, SGP offers significant efficiency in both memory and time complexity.
Note that GPs do not fit in our 80GB GPU memory, and we exclude them from baselines.
Fusion utilizes both LiDAR and image encoders and employs a single inpainting network to predict elevation.
Please refer to the supplementary materials for details of other baselines.

\begin{table}[t!]
\begin{center}
\footnotesize
\scalebox{0.7}{
\begin{tabular}{c|c|ccc|ccc|ccc}
\toprule
\multirow{2}{*}{Dataset} & \multirow{2}{*}{Method} & \multicolumn{3}{c|}{Elevation Err. (m) $\downarrow$} & \multicolumn{3}{c|}{Slope Err. (\%) $\downarrow$} & \multicolumn{3}{c}{Curvature Err. (1/m) $\downarrow$} \\
&& Total & Obs. & Unobs. & Total & Obs. & Unobs. & Total & Obs. & Unobs. \\
\drule
\multirow{12}{*}{CA Hills} & SGP & 1.825 & 1.029 & 2.062 & 5.3 & 4.4 & 5.5 & \textbf{0.036} & 0.034 & \textbf{0.036}\\
& DeepGP & 1.850 & 1.047 & 2.091 & 10.0 & 6.8 & 11.0 & 0.065 & 0.048 & 0.071 \\
& HMs + RFFs & 1.795 & 0.990 & 2.036 & 5.6 & 4.3 & 6.0 & 0.039 & 0.035 & 0.040 \\
& BEVNet~\cite{bevnet} & 0.591 & \textbf{0.274} & 0.686 & 4.7 & 3.2 & 5.2 & 0.041 & 0.030 & 0.045 \\
& TerrainNet~\cite{terrainnet} &  0.806 & - & - & 5.1 & - & - & 0.041 & - & -\\
& Fusion~\cite{bevnet, terrainnet} & 0.628 & 0.287 & 0.730 & 4.8 & 3.3 & 5.3 & 0.042 & 0.032 & 0.046 \\
& \textbf{Ours} & \textbf{0.586} & 0.280 & \textbf{0.677} & \textbf{4.2} & \textbf{3.1} & \textbf{4.6} & \textbf{0.036} & \textbf{0.028} & 0.039 \\
\cmidrule{2-11}
& HMs + RFFs-TA & 1.191 & 0.672 & 1.934 & 4.8 & 4.2 & 5.8 & \textbf{0.036} & 0.034 & \textbf{0.039} \\
& BEVNet-TA~\cite{bevnet} & 0.570 & 0.276 & 0.659 & 4.8 & 3.2 & 5.2 & 0.041 & \textbf{0.031} & 0.045\\
& TerrainNet-TA~\cite{terrainnet} & 0.791 & - & - & 5.3 & - & - & 0.045 & - & -\\
& Fusion-TA~\cite{bevnet, terrainnet} & 0.527 & \textbf{0.272} & 0.604 & 4.8 & 3.5 & 5.3 & 0.045 & 0.034 & 0.048 \\
& \textbf{Ours-TA} & \textbf{0.512} & 0.304 & \textbf{0.574} & \textbf{4.2} & \textbf{3.3} & \textbf{4.5} & 0.038 & \textbf{0.031} & 0.041 \\
\midrule
\multirow{12}{*}{Mojave} & SGP & 0.733 & 0.409 & 0.808 & 3.2 & 3.4 & 3.1 & 0.027 & 0.030 & 0.026 \\
& DeepGP & 0.765 & 0.427 & 0.844 & 7.2 & 5.7 & 7.6 & 0.050 & 0.042 & 0.053 \\
& HMs + RFFs & 0.678 & 0.334 & 0.758 & 3.0 & 2.9 & 3.0 & 0.027 & 0.028 & 0.027\\
& BEVNet~\cite{bevnet} & 0.265 & 0.141 & 0.294 & 2.8 & 2.3 & 2.9 & 0.027 & 0.022 & 0.029 \\
& TerrainNet~\cite{terrainnet} & 0.455 & - & - & 3.1 & - & - & 0.028 & - & - \\
& Fusion~\cite{bevnet, terrainnet} & 0.271 & 0.141 & 0.302 & 2.8 & 2.3 & 3.0 & 0.028 & 0.024 & 0.030\\
& \textbf{Ours} & \textbf{0.255} & \textbf{0.131} & \textbf{0.284} & \textbf{2.5} & \textbf{2.0} & \textbf{2.6} & \textbf{0.023} & \textbf{0.019} & \textbf{0.025} \\
\cmidrule{2-11}
& HMs + RFFs-TA & 0.452 & 0.222 & 0.779 & 2.8 & 2.6 & 3.1 & 0.026 & 0.026 & 0.026\\
& BEVNet-TA~\cite{bevnet} & 0.354 & 0.162 & 0.399 & 3.0 & 2.4 & 3.2 & 0.030 & 0.025 & 0.031\\
& TerrainNet-TA~\cite{terrainnet} & 0.452 & - & - & 3.3 & - & -& 0.030 & - & -\\
& Fusion-TA~\cite{bevnet, terrainnet} & 0.229 & \textbf{0.137} & 0.251 & 3.0 & 2.4 & 3.1 & 0.031 & 0.026 & 0.033 \\
& \textbf{Ours-TA} & \textbf{0.225} & 0.154 & \textbf{0.242} & \textbf{2.4} & \textbf{2.1} & \textbf{2.5} & \textbf{0.023} & \textbf{0.020} & \textbf{0.024} \\
\midrule
\multirow{12}{*}{Ellensburg} & SGP & 1.232 & 0.591 & 1.373 & 5.4 & 4.2 & 5.7 & \textbf{0.039} & 0.035 & \textbf{0.041} \\
& DeepGP & 1.194 & 0.583 & 1.328 & 8.6 & 6.2 & 9.2 & 0.059 & 0.047 & 0.063 \\
& HMs + RFFs & 1.085 & 0.541 & 1.205 & 5.3 & 3.9 & 5.6 & 0.040 & 0.035 & \textbf{0.041} \\
& BEVNet~\cite{bevnet} & 0.636 & 0.298 & 0.710 & 5.7 & 3.8 & 6.1 & 0.049 & 0.036 & 0.053\\
& TerrainNet~\cite{terrainnet} &  0.730 & - & - & 5.4 & - & - & 0.043 & - & - \\
& Fusion~\cite{bevnet, terrainnet} & 0.697 & 0.307 & 0.782 & 5.9 & 3.9 & 6.3 & 0.052 & 0.037 & 0.057\\
& \textbf{Ours} & \textbf{0.585} & \textbf{0.282} & \textbf{0.652} & \textbf{5.0} & \textbf{3.6} & \textbf{5.4} & 0.041 & \textbf{0.033} & 0.043 \\
\cmidrule{2-11}
& HMs + RFFs-TA & 0.783 & 0.443 & 1.132 & 5.1 & 4.3 & 5.9 & \textbf{0.039} & 0.037 & \textbf{0.043} \\
& BEVNet-TA~\cite{bevnet} & 0.693 & 0.316 & 0.776 & 5.7 & 4.0 & 6.1 & 0.049 & 0.037 & 0.052 \\
& TerrainNet-TA~\cite{terrainnet} & 0.746 & - & - & 5.6 & - & - & 0.044 & - & -\\
& Fusion-TA~\cite{bevnet, terrainnet} & 0.640 & \textbf{0.298} & 0.715 & 5.8 & 4.0 & 6.2 & 0.052 & 0.039 & 0.056 \\
& \textbf{Ours-TA} & \textbf{0.586} & 0.299 & \textbf{0.649} & \textbf{5.2} & \textbf{3.7} & \textbf{5.5} & 0.044 & \textbf{0.034} & 0.046 \\
\bottomrule
\end{tabular}
}
\end{center}
\vspace{-0.1cm}
\caption{\textbf{Comparisons with baseline on the CA Hills, Mojave deserts, and Ellensburg sequences.}}
\vspace{-0.7cm}
\label{tab:total}
\end{table}

\begin{table}[t!]
\begin{center}
\footnotesize
\scalebox{0.7}{
\begin{tabular}{c|c|c|c|c}
\toprule
Dataset & Method & Elev. (Gen. Err.) & Slope (Gen. Err.) & Curv. (Gen. Err.)\\
\drule
\multirow{5}{*}{\shortstack{Ellensburg\\$\rightarrow$CA Hills}} & SGP & 2.853 \textcolor{magenta}{(+0.028)} & 9.0 \textcolor{magenta}{(+3.7)} & 0.082 \textcolor{magenta}{(+0.046)} \\
&BEVNet~\cite{bevnet} & 1.251 \textcolor{magenta}{(+0.660)}  & 6.4 \textcolor{magenta}{(+1.7)} & 0.052 \textcolor{magenta}{(+0.011)} \\
& TerrainNet~\cite{terrainnet} & 1.346 \textcolor{magenta}{(+0.540)} & 6.0 \textcolor{magenta}{(+0.9)} & \textbf{0.043 \textcolor{magenta}{(+0.002)}} \\
& Fusion~\cite{bevnet, terrainnet} & 1.185 \textcolor{magenta}{(+0.557)} & 6.0 \textcolor{magenta}{(+1.2)} & 0.050 \textcolor{magenta}{(+0.008)} \\
& \textbf{Ours} & \textbf{1.113 \textcolor{magenta}{(+0.527)}} & \textbf{5.4 \textcolor{magenta}{(+1.2)}} & 0.044 \textcolor{magenta}{(+0.008)}\\
\midrule
\multirow{5}{*}{\shortstack{Ellensburg\\$\rightarrow$Mojave}} & SGP & 1.775 \textcolor{magenta}{(+0.042)} & 4.4 \textcolor{magenta}{(+1.2)} & 0.043 \textcolor{magenta}{(+0.016)} \\
& BEVNet~\cite{bevnet} & 0.542 \textcolor{magenta}{(+0.277)} & 3.7 \textcolor{magenta}{(+0.9)} & 0.037 \textcolor{magenta}{(+0.010)} \\
& TerrainNet~\cite{terrainnet} & 0.578 \textcolor{magenta}{(+0.123)} & 3.3 \textcolor{magenta}{(+0.2)} & 0.029 \textcolor{magenta}{(+0.001)} \\
& Fusion~\cite{bevnet, terrainnet} & 0.599 \textcolor{magenta}{(+0.328)} & 3.6 \textcolor{magenta}{(+0.8)} & 0.038 \textcolor{magenta}{(+0.010)} \\
& \textbf{Ours} & \textbf{0.446 \textcolor{magenta}{(+0.191)}} & \textbf{3.0 \textcolor{magenta}{(+0.5)}} & \textbf{0.028 \textcolor{magenta}{(+0.005)}} \\
\bottomrule
\end{tabular}
}
\end{center}
\caption{\textbf{Comparisons with baselines under two generalization scenarios}: Ellensburg$\rightarrow$CA Hills and Ellensburg$\rightarrow$Mojave Desert. Magenta errors are the gaps between in-domain and generalization.}
\vspace{-1.0cm}
\label{tab:generalization}
\end{table}

\vspace{-0.4cm}
\subsection{Quantitative Comparisons}
\vspace{-0.3cm}
\noindent\textbf{In-domain Comparisons.} Table~\ref{tab:total} presents comparisons with baselines on three sequences in the entire, observed, and unobserved areas.
Our method achieves the lowest errors on most metrics across all sequences.
Importantly, our method consistently achieves the lowest slope errors.
In high-speed off-road navigation, precisely estimating slopes is essential not only to maintain high speed on various terrains but also to avoid crashes and rollovers.
BEVNet~\cite{bevnet} achieves the second-lowest errors in most cases, demonstrating the effectiveness of using LiDAR points for predicting elevations.
On the other hand, the pure image-based method, TerrainNet~\cite{terrainnet}, fails to achieve comparable accuracy in most cases.
This is because images provide indirect semantic information, whereas LiDAR points provide direct supervision of ground heights.
Another interesting observation is that our method outperforms the Fusion method~\cite {bevnet, terrainnet} by a large margin, especially in terms of slope and curvature.
While both methods utilize the same modalities, Fusion even fails to outperform BEVNet~\cite{bevnet}, proving the difficulty of fusing different modalities.

Temporally aggregating observations in our method significantly improves elevation errors in the CA Hills and Mojave Desert sequences.
However, we observe two interesting points: 1) Applying aggregation to Ellensburg sequences does not improve much, and 2) errors on observed regions at each time $t$ increase after aggregation.
We conjecture that the first issue occurs because the vehicle's speed was slower in Ellensburg than in the other sequences, due to the presence of many negative obstacles.
That is, the observation space is still limited even after aggregation, failing to improve as much as in the other sequences.
Indeed, if we compare the error gain of elevation errors in unobserved regions, the gain in Ellensburg is only 0.003, while it is 0.103 and 0.042 in CA Hills and Mojave Deserts.

For the second issue, we conjecture that incorrect handling of observation uncertainty in LiDAR readings may lead to a performance drop in observed regions.
In LiDAR aggregation, we do not incorporate their associated uncertainty, e.g., degradation of measurement confidence over time, from the past observations.
Instead, they have the same impact as the current observation.
This problem can be mitigated by associating it with the corresponding uncertainty, but this remains future work.

\noindent\textbf{Generalization.} Generalization of methods to unseen environments is a pivotal concern for deploying robots on various sites.
To demonstrate their generalization capability, we evaluate the methods under two scenarios: from Ellensburg to the CA Hills and from Ellensburg to the Mojave Desert.
As shown in Table~\ref{tab:generalization}, our method exhibits the lowest errors on most metrics and scenarios.
More importantly, our method presents the lowest generalization error gaps in elevation errors (magenta-colored errors).
That is, ours achieves the best elevation accuracy and also the lowest generalization gap among baselines.
On the other hand, BEVNet~\cite{bevnet} suffers from huge generalization errors in most metrics.

\noindent\textbf{Uncertainty.}
To evaluate the quality of predicted uncertainty, we modify the BEVNet training to estimate predictive distribution by changing the loss function from Smooth-L1 to negative log-likelihood (NLL).
We report NLL and expected normalized calibration error (ENCE)~\cite{ence} for comparisons.
As shown in Table~\ref{tab:uncert_ablation} (left), uncertainty training of BEVNet accompanies significant accuracy drops in elevation predictions on all three sequences. Furthermore, in the case of Ellensburg, we observe very high NLL values during BEVNet training, indicating that the predicted distribution diverges strongly from the ground-truth elevation.
On the other hand, our method achieves reasonable NLL and ENCE values, outperforming BEVNet on Mojave and Ellensburg sequences. 
However, our method fails to surpass SGP on both metrics. This is because NPs present estimated, amortized uncertainty from learned networks, whereas SGP computes uncertainty through direct Bayesian inference conditioned on observed data.

\vspace{-0.4cm}
\subsection{Qualitative Comparisons}
\begin{figure}[t!]
  \centering
  \vspace{-0.3cm}
  \includegraphics[width=0.8\textwidth]{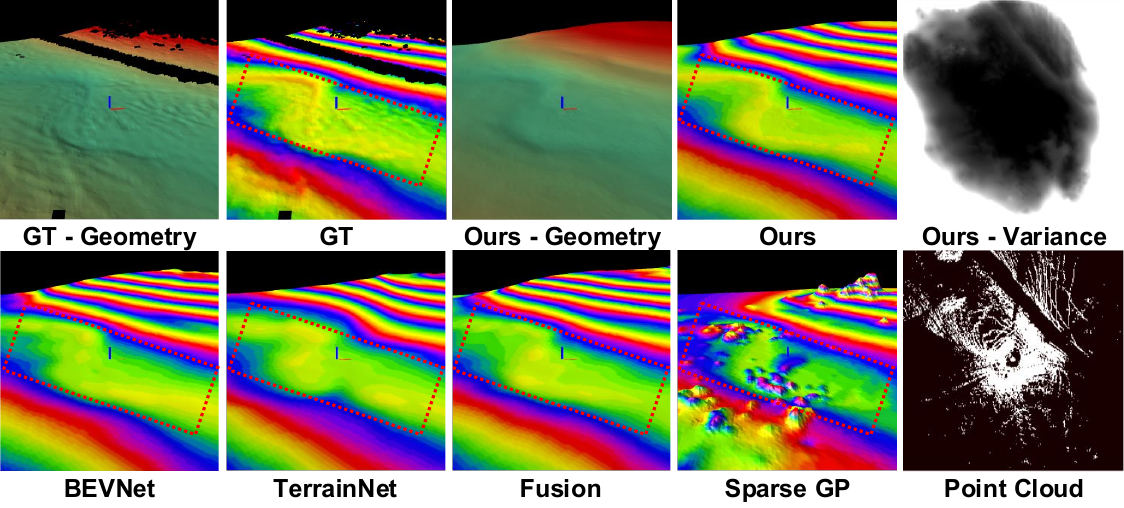}
  \vspace{-0.3cm}
  \caption{\textbf{Qualitative comparisons on the CA Hills sequence.} The predicted elevations are visualized in 3D and also color-coded with a repeated pattern.}
  \vspace{-0.4cm}
\label{fig:qualitative}
\end{figure}

\vspace{-0.3cm}
Fig.~\ref{fig:qualitative} illustrates the ground truth and elevation predictions from our method and baselines.
We visualize 3D elevations and their colored meshes with repeated patterns to observe height differences.
Also, we visualize the predicted uncertainty (i.e., variance) from our method with the point cloud.
As shown, our method accurately estimates narrow ditches, while other baselines fail to capture them.
Moreover, our uncertainty predictions align with the observations from the point cloud.
More qualitative examples can be found in the supplementary materials.

\vspace{-0.3cm}
\subsection{Ablation Studies}
\vspace{-0.3cm}
\noindent\textbf{Effectiveness of Each Modality.}
The upper group of Table~\ref{tab:uncert_ablation} (right) demonstrates the effectiveness of using LiDAR and/or image modalities for elevation predictions. As reported, our method benefits from both modalities significantly.
Also, considering that SGP and our no-semantic method take the same inputs, NPs present a more effective, flexible modeling of terrains.

\begin{table}[t!]
\begin{center}
\footnotesize
\begin{subtable}[t]{0.48\linewidth}
\centering
\scalebox{0.73}{
    \begin{tabular}{c|c|ccccc}
    \toprule
    Dataset & Method & Elev. & Slope & Curv. & NLL$_\downarrow$ & ENCE$_\downarrow$ \\
    \drule
    \multirow{3}{*}{\shortstack{Ellensburg\\$\rightarrow$CA Hills}} & SGP & 1.853 & 9.0 & 0.082 & \textbf{4.58} & \textbf{1.76} \\
    & BEVNet-NLL & 1.203 & 8.2 & 0.084& 16801.92 & 3.90\\
    & Ours & \textbf{1.113} & \textbf{5.4} & \textbf{0.044} & 8.46 & 2.20\\
    \midrule
    \multirow{3}{*}{\shortstack{Ellensburg\\$\rightarrow$Mojave}} & SGP & 0.775 & 4.4 & 0.043 & \textbf{0.61} & \textbf{0.44} \\
    & BEVNet-NLL & \textbf{0.415} & 3.4 & 0.034 & 2.12 & 0.86 \\
    & Ours & 0.446 & \textbf{3.0} & \textbf{0.028} & 1.37 & 0.63 \\
    \bottomrule
    \end{tabular}}
    \vspace{-0.2cm}
\end{subtable}
\hfill
\begin{subtable}[t]{0.48\linewidth}
\centering
\scalebox{0.73}{
    \begin{tabular}{l|ccc}
    \toprule
    Method & Elev. & Slope & Curv.  \\
    \drule
    No Semantics & 1.566 & 9.6 & 0.099 \\
    + LiDAR & 0.595 & \textbf{4.2} & \textbf{0.035} \\
    \textbf{+ Image} & \textbf{0.586} & \textbf{4.2} & 0.036 \\
    \midrule
    Single & 0.586 &  4.2 & \textbf{0.036} \\
    +LiDAR Agg. & 0.513  & 4.2 & 0.038 \\
    \textbf{+Img. Bayes. Update} & \textbf{0.512} & 4.2 & 0.038  \\
    \bottomrule
    \end{tabular}}
    \vspace{-0.1cm}
\end{subtable}
\end{center}
\caption{\textbf{Uncertainty comparisons and ablation studies.} (left): Comparisons of uncertainty prediction between ours and BEVNet trained with negative log-likelihood (NLL) loss. (right): Ablation studies of our method. Upper group: Effectiveness of each modality. Lower group: Effectiveness of temporal aggregation types.}
\vspace{-0.8cm}
\label{tab:uncert_ablation}
\end{table}

\noindent\textbf{Effectiveness of Aggregations.}
The lower group of Table~\ref{tab:uncert_ablation} (right) shows the effectiveness of aggregations of each modality.
Using aggregated LiDAR points improves performance significantly, while learning-based aggregation~\cite{bevnet, terrainnet} sometimes fails to improve, as shown in Table~\ref{tab:total}.
While adding image aggregation brings improvements in elevation predictions, it is not significant.
This is because images only provide indirect supervision to elevations, whereas LiDAR provides more direct information.

\vspace{-0.4cm}
\section{Conclusion}
\vspace{-0.3cm}
In this paper, we demonstrate the potential of using Neural Processes for terrain elevation modeling in off-road environments.
We utilize semantic information from LiDAR and camera sensors to condition the context and target points in Neural Processes.
Also, we effectively reduce floating-point operations with ball-query attentions while ensuring spatially consistent and precise elevation predictions.
Our method alleviates the elevation over-smoothing tendencies of existing methods and provides predictive uncertainty.
We validated our approach on three different off-road sequences collected from different testing sites and demonstrated that our method outperforms competing elevation estimation methods.

\section{Limitation}
While our method demonstrates strong performance, there remain limitations for future work.
First, our temporal aggregation can be further improved by compensating for the ego vehicle's elevation difference and leveraging the learned aggregation module.
While incorporating the ego vehicle's elevation into learned modules is not trivial, we believe there is room to adopt a more capable learned aggregation module for both LiDAR and image modalities.
Secondly, developing representations with adaptive resolution is another future work.
We assume a regular grid of BEV maps to predict elevations, but Neural Processes learn continuous functions.
Thus, we can adaptively sample the elevation if we need a finer-grained resolution in a particular region, and vice versa.
While this is still possible for us, our downstream planning and control algorithms are still locked in fixed BEV grids.
Devising an advanced representation will be our future work.
Additionally, as presented in uncertainty comparisons, NPs predict uncertainty from learned networks; therefore, they must be appropriately calibrated.
Lastly, we have not validated our method beyond our desired ranges, i.e., -51.2 to 51.2 meters.
Considering that stochastic processes often fail to extrapolate context points to target points far from them, Neural Processes may also have such shortcomings.
Identifying and analyzing such problems will be our future work.
\section{Acknowledgement}
This research was developed with funding from the Defense Advanced Research Projects Agency (DARPA).




\clearpage

\appendix
\section{Bayesian Filtering of Image BEV Features}
This section details how we temporally aggregate image BEV features using Bayesian updates.
As aforesaid in the manuscript, we lift and splat the image features at time $t$ to the BEV space and use the normalized density of each BEV grid as the probability of being correct, $\hat{\mathbf{p}}_t$.
Then, the temporally aggregated image features $\mathbf{f}_t$ are obtained by
\begin{gather}
   \mathbf{f}_t = \frac{\hat{\mathbf{p}}_t\,\hat{\mathbf{f}}_t + \mathbf{p}_{t-1} \mathbf{f}_{t-1}}{\hat{\mathbf{p}}_t + \mathbf{p}_{t-1}}, \ \ \ \   
   \mathbf{p}_t = \frac{\hat{\mathbf{p}}_t\mathbf{p}_{t-1}}{\hat{\mathbf{p}}_t\mathbf{p}_{t-1} + (1 - \hat{\mathbf{p}}_t)(1 - \mathbf{p}_{t-1})},  
\end{gather}
where $\hat{\mathbf{f}}_t$ denote the image features at time $t$ before aggregation, and $\mathbf{p}_t$ indicate the aggregated probability.

\noindent\textbf{Lifting Image Features.} With given image intrinsic matrix $\mathbf{K}\in\mathbb{R}^{3\times3}$ and extrinsic matrix of camera to ego vehicle's base $[\mathbf{R}|\mathbf{t}]\in\mathbb{R}^{3\times4}$, each pixel location $u, v$ with its stereo depth $d$ is lifted to the ego vehicle's space by
\begin{gather}
   \begin{pmatrix} x\\y\\z \end{pmatrix} = d\begin{pmatrix} u \\ v \\ 1 \end{pmatrix} (\mathbf{K}^{-1})^\intercal \mathbf{R}^\intercal + \mathbf{t}.
\end{gather}

\noindent\textbf{Splatting.}
Afterward, $x, y$ values are truncated within predefined ranges, i.e., $x$: [-51.2 m, 51.2 m] and $y$: [-51.2 m, 51.2 m], and binned into the BEV grid using a resolution of $R=0.4$ m.
For each BEV cell, we average the lifted image features (i.e., $\hat{\mathbf{f}}_{t}^{x,y}$) and also count the number of image features (i.e., density $\sigma^{x,y}_t$).
We normalize the density using the maximum density at time $t$ to have a probability of each cell for being correct, $\hat{\mathbf{p}}_t$.

\noindent\textbf{Temporal Aggregation of Probabilities.}
Probability $\hat{\mathbf{p}}_t$ is the probability of being correct (i.e., $\text{P}(C))$ for each cell with given observations $\hat{\mathbf{f}}_t$, i.e., $\hat{\mathbf{p}}_t = \text{P}(C|\hat{\mathbf{f}}_t)$.
Accordingly, $\mathbf{p}_{t-1} = \text{P}(C|\mathbf{f}_{t-1})$ indicates the probability of being correct for previously aggregated features.
Then, we aim to obtain $\mathbf{p}_t = \text{P}(C|\hat{\mathbf{f}}_t, \mathbf{f}_{t-1}) = \text{P}(C|\mathbf{f}_t)$.
Using Bayes' rule, 
\begin{gather}
\text{P}(C|\hat{\mathbf{f}}_t, \mathbf{f}_{t-1}) = \frac{\text{P}(\hat{\mathbf{f}}_t, \mathbf{f}_{t-1}|C)\text{P}(C)}{\text{P}(\hat{\mathbf{f}}_t, \mathbf{f}_{t-1})}.\nonumber
\end{gather}
Under the conditional independence assumption, i.e., $\text{P}(\hat{\mathbf{f}}_t, \mathbf{f}_{t-1}|C) = \text{P}(\hat{\mathbf{f}}_t|C)\text{P}(\mathbf{f}_{t-1}|C)$, we can obtain the following:
\begin{align}
    \text{P}(C|\hat{\mathbf{f}}_t, \mathbf{f}_{t-1}) &= \frac{\text{P}(\hat{\mathbf{f}}_t|C)\text{P}(\mathbf{f}_{t-1}|C)\text{P}(C)}{\text{P}(\hat{\mathbf{f}}_t, \mathbf{f}_{t-1})} \nonumber \\
    \text{P}(\neg C|\hat{\mathbf{f}}_t, \mathbf{f}_{t-1}) &= \frac{\text{P}(\hat{\mathbf{f}}_t|\neg C)\text{P}(\mathbf{f}_{t-1}|\neg C)\text{P}(\neg C)}{\text{P}(\hat{\mathbf{f}}_t, \mathbf{f}_{t-1})}.
    \label{eq:cond_independence}
\end{align}
We can relate each term, $\text{P}(\hat{\mathbf{f}}_t|C), \text{P}(\mathbf{f}_{t-1}|C)$ to $\hat{\mathbf{p}}_t, \mathbf{p}_{t-1}$ using Bayes' rule:
\begin{align}
    \text{P}(\hat{\mathbf{f}}_t|C) &= \frac{\text{P}(C|\hat{\mathbf{f}}_t) \text{P}(\hat{\mathbf{f}}_t)}{\text{P}(C)} = \frac{\hat{\mathbf{p}}_t \text{P}(\hat{\mathbf{f}}_t)}{\text{P}(C)} \nonumber \\
    \text{P}(\mathbf{f}_{t-1}|C) &= \frac{\text{P}(C|\mathbf{f}_{t-1}) \text{P}(\mathbf{f}_{t-1})}{\text{P}(C)} = \frac{\mathbf{p}_{t-1}\text{P}(\mathbf{f}_{t-1})}{\text{P}(C)}. \nonumber
\end{align}
From Equ.~\eqref{eq:cond_independence}, we can obtain
\begin{equation}
    \text{P}(C|\hat{\mathbf{f}}_t, \mathbf{f}_{t-1}) = \frac{\hat{\mathbf{p}}_t \text{P}(\hat{\mathbf{f}}_t)}{\text{P}(C)} \cdot \frac{\mathbf{p}_{t-1}\text{P}(\mathbf{f}_{t-1})}{\text{P}(C)} \cdot \frac{\text{P}(C)}{\text{P}(\hat{\mathbf{f}}_t, \mathbf{f}_{t-1})} = \frac{\hat{\mathbf{p}}_t\mathbf{p}_{t-1} \text{P}(\hat{\mathbf{f}}_t)\text{P}(\mathbf{f}_{t-1})}{\text{P}(C)\text{P}(\hat{\mathbf{f}}_t, \mathbf{f}_{t-1})}.\nonumber
\end{equation}
Similarly,
\begin{equation}
    \text{P}(\neg C|\hat{\mathbf{f}}_t, \mathbf{f}_{t-1}) =  \frac{(1 - \hat{\mathbf{p}}_t)(1 - \mathbf{p}_{t-1}) \text{P}(\hat{\mathbf{f}}_t)\text{P}(\mathbf{f}_{t-1})}{\text{P}(\neg C)\text{P}(\hat{\mathbf{f}}_t, \mathbf{f}_{t-1})}.\nonumber
\end{equation}
Then, $\mathbf{p}_t = \text{P}(C|\hat{\mathbf{f}}_t, \mathbf{f}_{t-1})$ is
\begin{align}
   \text{P}(C|\hat{\mathbf{f}}_t, \mathbf{f}_{t-1}) &= \frac{\text{P}(C|\hat{\mathbf{f}}_t, \mathbf{f}_{t-1})}{\text{P}(C|\hat{\mathbf{f}}_t, \mathbf{f}_{t-1}) + \text{P}(\neg C|\hat{\mathbf{f}}_t, \mathbf{f}_{t-1})}  \nonumber \\
   &= \frac{\frac{\hat{\mathbf{p}}_t\mathbf{p}_{t-1}}{\text{P}(C)}}{\frac{\hat{\mathbf{p}}_t\mathbf{p}_{t-1}}{\text{P}(C)} + \frac{(1 - \hat{\mathbf{p}}_t)(1 - \mathbf{p}_{t-1})}{\text{P}(\neg C)}}.
   \label{eq:normalize}
\end{align}
Assuming we have a uniform prior, $\text{P}(C) = \text{P}(\neg C) = 0.5$, Equ.~\eqref{eq:normalize} reduces to
\begin{equation}
    \mathbf{p}_t = \text{P}(C|\hat{\mathbf{f}}_t, \mathbf{f}_{t-1}) = \frac{\hat{\mathbf{p}}_t\mathbf{p}_{t-1}}{\hat{\mathbf{p}}_t\mathbf{p}_{t-1} + (1 - \hat{\mathbf{p}}_t)(1 - \mathbf{p}_{t-1})}.
\end{equation}

\section{Related Work}
\subsection{Elevation Modeling}
Deep learning methods~\cite{elevation2_tal, terrainnet, bevnet, elevation3, elevation5, vstrong} have recently emerged as powerful alternatives for GPs and BGKs due to their strong capabilities in learning. TerrainNet~\cite{terrainnet} segments out semantics on BEV space and predicts their elevations from camera sensors. Chung \textit{et al.}~\cite{pixel_to_elev} also predicts elevations from camera sensors but in a longer range, covering around 100 m$\times$100 m range with transformer-based architecture. We adopt TerrainNet~\cite{terrainnet} as one of the baselines for comparisons. We also adopt BEVNet~\cite{bevnet} as a baseline, which estimates traversability using LiDAR readings in both urban scenes and off-road environments. As the authors noted in their paper, BEVNet can be easily extended to the elevation modeling task. Additionally, we merge both methods and use a fused model as another strong baseline.

\subsection{Neural Processes}
NPs are a family of meta-learning models that combine the flexibility of neural networks with the probabilistic advantages of stochastic processes~\cite{np, np_survey}.
NPs~\cite{anp, tnp, bnp, anp} learn data-driven representations without fixed kernel assumptions, enabling flexible predictions and efficient inference. 
Early NP variants suffered from context aggregation bottlenecks, causing overly smooth predictions. 
To address this, Attentive Neural Processes (ANPs) introduced cross-attention between context and target points, significantly improving predictive sharpness and uncertainty modeling~\cite{anp,convcnp}. However, incorporating semantic context and localized attention for sparse spatial data, particularly crucial in robotics applications, remains relatively unexplored.

\subsection{Occupancy Mapping}
Our task and occupancy mapping~\cite{hilbert_maps, bayesian_hilbert_maps, GP_OM, diffusion_OM} share conceptual foundations in modeling spatially continuous environments under uncertainty. Hilbert Maps (HMs)~\cite{hilbert_maps} approximate continuous, binary occupancy based on kernel approximations such as random Fourier features.
Bayesian Hilbert Maps (BHMs)~\cite{bayesian_hilbert_maps} extend this to learn long-term occupancy using a variational Bayesian approach.
More recent works further enhance occupancy mapping through Gaussian process field shaping~\cite{GP_OM} and generative modeling techniques, such as diffusion-based synthesis~\cite{diffusion_OM}. However, key differences remain.
Occupancy mapping typically models binary 3D occupancy, with emphasis on reconstructing observed regions or estimating traversable free space.
On the other hand, our work focuses on terrain elevation modeling, a 2.5D surface representation, which requires not only interpolating sparse observations but also extrapolating to unobserved areas.
Thus, learning expressive spatial priors that can capture fine-grained geometry, such as negative obstacles or sudden slopes, is essential.
Additionally, our approach necessitates learning to remove non-ground structures (e.g., vegetation, vehicles), a challenge not typically addressed in standard occupancy mapping.

\section{Additional Experimental Setups and Results}
This section provides details about the baselines and implementation, then presents additional ablation studies and qualitative comparisons.

\subsection{Baselines and Implementation Details}
We adopt Deep Gaussian Processes (DeepGPs)~\cite{deepgp}, Hilbert Maps (HMs) with random Fourier features (RFFs)~\cite{hilbert_maps}, Sparse Gaussian Processes (SGP)~\cite{sgp1, sgp2, sgp3}, BEVNet~\cite{bevnet}, TerrainNet~\cite{terrainnet}, and Fusion~\cite{bevnet, terrainnet} as our baselines.

\noindent\textbf{DeepGPs.} We adopt two-layered DeepGPs with 8 hidden dimensions. We use a RBF kernel with 128 and 64 inducing points for the first and last layers. Implementations are based on  the GPytorch~\cite{gpytorch} library.

\noindent\textbf{Hilbert Maps with RFFs.} We adopt the offical implementation of Hilbert Maps~\cite{hilbert_maps} and modify the binary classifier with a MLP-based regression.
We use the number of Fourier features of 1,000 and $\gamma=1.0$.

\noindent\textbf{SGP.} We use a Rational Quadratic kernel with $\Theta = 0.7\mathbf{I}$ and $\alpha=10$:
\begin{equation*}
  k_{\text{RQ}}(\mathbf{x_1}, \mathbf{x_2}) =  \left(1 + \frac{1}{2\alpha}
  (\mathbf{x_1} - \mathbf{x_2})^\top \Theta^{-2} (\mathbf{x_1} - \mathbf{x_2}) \right)^{-\alpha}.
\end{equation*}
We set the maximum number of inducing points to 14,000.
The above parameters are found by optimizing hyperparameters to maximize the marginal log likelihood on a subset of validation samples. 

\noindent\textbf{BEVNet.} We modify the original implementation to have a front-facing image and add an image encoder (i.e., EfficientNet-B0~\cite{efficientnet}).
The predicted image features are concatenated to the corresponding LiDAR points.
This is to improve the prediction accuracy on the vehicle's front regions.
We train the model using a SmoothL1 loss function with $\beta=0.2$.
The model parameters are optimized using the Adam optimizer~\cite{adam} with a step learning rate scheduler.
Note that we use the same optimizer and learning rate scheduler across all baselines with a learning rate of 1e-3.

\noindent\textbf{TerrainNet.} We follow the original implementation of TerrainNet~\cite{terrainnet}.
We first train the depth completion model across all dataset sequences using the cross-entropy loss function.
Then, we train the full model using the Smooth L1 loss for elevation predictions and cross-entropy loss for depth completion.
Originally, TerrainNet was validated in the range of [-38.4 m, 38.4 m] because stereo depth and completed depth accuracy drop significantly in long ranges.
However, we adopt [-51.2 m, 51.2 m] range for fair comparisons with other baselines.
The results can be improved by adopting a shorter range for TerrainNet.

\noindent\textbf{Fusion.} The fused model of BEVNet and TerrainNet has an identical LiDAR encoder and inpainting module.
To incorporate internet-scale knowledge, we replace the image encoder from EfficientNet-B0 to Dino-V2~\cite{dinov2}.
For recurrency, we aggregate LiDAR points, identical to our method, but train ConvGRU as proposed in BEVNet~\cite{bevnet} and TerrainNet~\cite{terrainnet}.
Identical to other baselines, it is trained with Smooth L1 loss.

\noindent\textbf{Our Method.}
We adopt $\epsilon = 2.0$ meters for the ball-query attention and set the maximum number of neighbors and the number of heads as 32 points and 4, respectively.
Also, we use the hidden dimension of 64 for both deterministic and latent paths.
During training, we set the maximum context points for each instance to 7,000, while we use the entire context points for validation.
We use a batch size of 4, and the model is trained with 4 NVIDIA A100 GPUs, 80 GB.

\subsection{Ablation Studies}
This section presents additional ablation studies about the effectiveness of ball query attention and computational cost.
\begin{table}[t!]
\begin{center}
\footnotesize
\begin{tabular}{c|ccc|ccc|ccc|c}
\toprule
& \multicolumn{3}{c|}{Elevation Err. (m) $\downarrow$} & \multicolumn{3}{c|}{Slope Err. (\%) $\downarrow$} & \multicolumn{3}{c}{Curvature Err. (1/m) $\downarrow$} & \\
Method & Total & Obs. & Unobs. & Total & Obs. & Unobs. & Total & Obs. & Unobs. & NLL \\
\drule
Global Attn. & 0.619 & 0.279 & 0.720 & \textbf{4.2} & 3.1 & \textbf{4.6} & \textbf{0.035} & 0.028 & \textbf{0.038} & 4.73 \\
BQA, $\epsilon=1.0$ & 0.597 & 0.277 & 0.693 & 4.3 & \textbf{3.0} & 4.7 & 0.036 & 0.028 & 0.039 & 3.25 \\
\textbf{BQA, $\mathbf{\epsilon=2.0}$} & \textbf{0.586} & 0.280 & \textbf{0.677} & \textbf{4.2} & 3.1 & \textbf{4.6} & 0.036 & 0.028 & 0.039 & \textbf{2.64} \\
BQA, $\epsilon=3.0$ & 0.603 & \textbf{0.274} & 0.701 & 4.3 & 3.1 & 4.7 & 0.036 & 0.028 & 0.039 & 3.63\\
\bottomrule
\end{tabular}
\end{center}
\vspace{-0.1cm}
\caption{\textbf{Ablation studies on global attention (i.e., Global Attn.) and ball query attention (i.e., BQA) across different $\epsilon$ values.}}
\vspace{-0.7cm}
\label{supptab:radius_ablation}
\end{table}

\noindent\textbf{Global Attention vs Ball Query Attention.}
We compare the ball query attention with global attention and demonstrate the effectiveness of different $\epsilon$ values in Table~\ref{supptab:radius_ablation}.
As shown, all kinds of attention effectively capture the slopes and curvature, while using global attention falls behind the ball query attention.
The error gap mainly comes from the errors in unobserved regions.
It shows an L1 error of 0.720, much higher than 0.677 from our best setting.
This could be because the nearest LiDAR points are much sparser in unobserved regions.
Therefore, if we apply global attention, the model loses locality in attention and is diluted by non-local information, failing to estimate elevations accurately.
On the other hand, ball query attention only leverages local information and effectively preserves local characteristics.
Among the ball query attention results, $\epsilon=2.0$ demonstrates the lowest elevation error and negative log-likelihood (NLL).
However, we observe no significant error gaps in slopes and curvatures.

\begin{table}[t!]
\begin{center}
\footnotesize
\begin{tabular}{c|ccccc}
\toprule
Method & Total Param. & Trainable Param. & GFLOPs & Memory & Infer. Time (Sec.) \\
\drule
BEVNet~\cite{bevnet} & 31.45M & 31.45M & \textbf{113.55} & 1,372MB & \textbf{0.025} \\
TerrainNet~\cite{terrainnet} & \textbf{18.57M} & \textbf{18.57M} & 241.05 & \textbf{1,112MB} & 0.053\\
Fusion~\cite{bevnet, terrainnet} & 41.99M & 19.93M & 295.31 & 2,286MB & 0.058 \\
Ours - BQA & 42.77M & 20.71M & 434.24 & 13,514MB & 0.161\\
Ours - Global Attn. &  42.77M & 20.71M & 521.66 & 60,812MB & 0.252\\
\bottomrule
\end{tabular}
\end{center}
\vspace{-0.1cm}
\caption{\textbf{Analysis of computational costs for each method.} GFLOPs and GPU memory are measured using the validation set of the CA Hills dataset, and the maximum GPU memory usage is reported throughout the sequence.}
\vspace{-0.7cm}
\label{supptab:computational_cost_ablation}
\end{table}
\begin{figure}[ht]
\centering
  \includegraphics[width=0.8\textwidth]{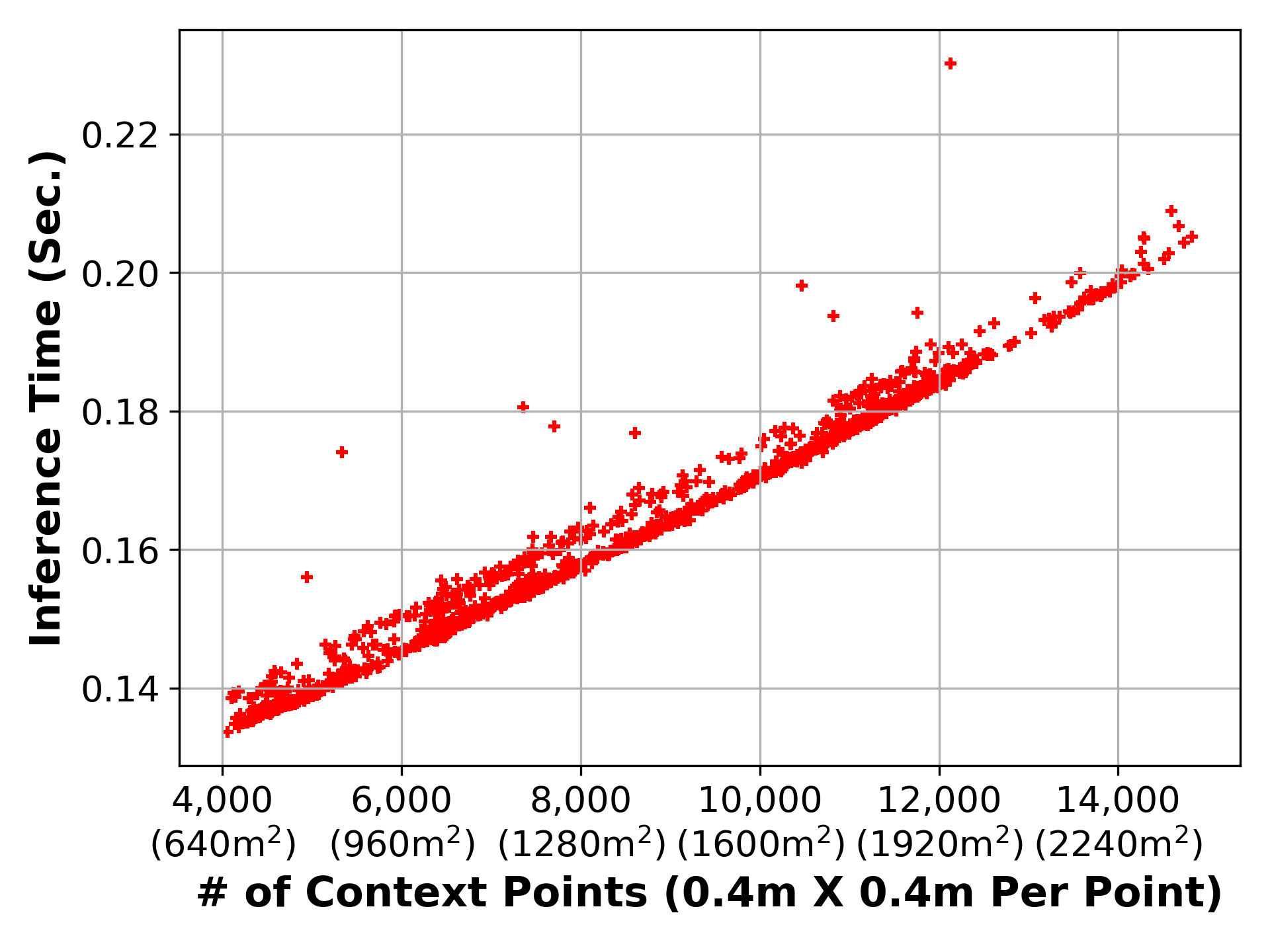}
\vspace{-0.3cm}
\caption{\textbf{An analysis of inference time for the number of context points.}}
\label{suppfig:infertime}
\end{figure}

\noindent\textbf{Computational Cost Analysis.}
Table~\ref{supptab:computational_cost_ablation} reports the total number of parameters, trainable parameters, floating operations, and GPU memory.
TerrainNet~\cite{terrainnet} contains the fewest parameters and GPU memory consumption, and BEVNet~\cite{bevnet} presents the fewest floating operations.
Fusion~\cite{bevnet, terrainnet} has around 40M parameters, requiring more GPU memory and FLOPS than other baselines.
In our methods, attention mechanisms require more floating operations and GPU memory.
Using global attention requires up to 60GB, which may not be feasible for onboard computing for off-road vehicles.
However, applying ball query attention significantly reduces GFLOPs and GPU memory consumption by 17\% and 78\%, respectively.
GPU memory usage can be further reduced by applying mixed precision (e.g., floating point 16), and it also allows faster inference speed.
Also, we further analyze our inference time by measuring inference time based on the number of context samples. Each context point covers 0.4m$\times$0.4m area, and as shown in Fig.~\ref{suppfig:infertime}, inference time linearly increases as more context points are used.

\noindent\textbf{Sample Efficiency.} We analyze the sample efficiency from two aspects: context points and training sample efficiency.
As shown in the tables (elev. error), our method maintains reasonable error rates even with small context sample sizes, thanks to semantic feature conditioning.
Additionally, our method maintains reasonable performance with varying numbers of randomly sampled training data points. We believe that our method is robust to sparse training samples, provided they are sampled uniformly at random across the entire dataset. 
In contrast, using the first n\% frames, the error more steeply degrades, experiencing greater distributional gaps.
Still, we believe our method maintains reasonable performance, without failing drastically.

\begin{table}[h]
\begin{center}
\footnotesize
\centering
\begin{tabular}{c|ccccc}
\toprule
& \multicolumn{5}{c}{\% of Observations as Context Points}\\
Ellensburg$\rightarrow$CA Hills & 20\% & 40\% & 60\% & 80\% & 100\% \\
\drule
Elevation Error& 1.136 & 1.123 & 1.123 & 1.115 & 1.113\\
\bottomrule
\end{tabular}
\end{center}
\caption{\textbf{Impact of the number of context points from observations on elevation accuracy.}}
\end{table}

\begin{table}[h]
\begin{center}
\footnotesize
\centering
\begin{tabular}{c|cccc}
\toprule
& \multicolumn{4}{c}{\% of training set} \\
CA Hills & 25\% & 50\% & 75\% & 100\% \\
\drule
Random samples & 0.679 & 0.604 & 0.599 & 0.586 \\
First n\% frames & 0.928 & 0.735 & 0.683 & 0.586 \\
\bottomrule
\end{tabular}
\end{center}
\caption{\textbf{Impact of the number of training samples on elevation accuracy.}}
\end{table}

\subsection{Qualitative Results.}
This section makes qualitative evaluations between ours and baselines, our single-frame and temporally-aggregated models, and different samples from varying global latent $z$ in our method.
This section presents three different qualitative evaluations. First, we compare our predictions with baselines.
And then, we demonstrate the effectiveness of using temporal aggregation compared to single-frame results.
Lastly, we show multiple elevation predictions from varying global latent $z$ of our method.

\begin{figure}[t!]
  \centering
  \includegraphics[width=\textwidth]{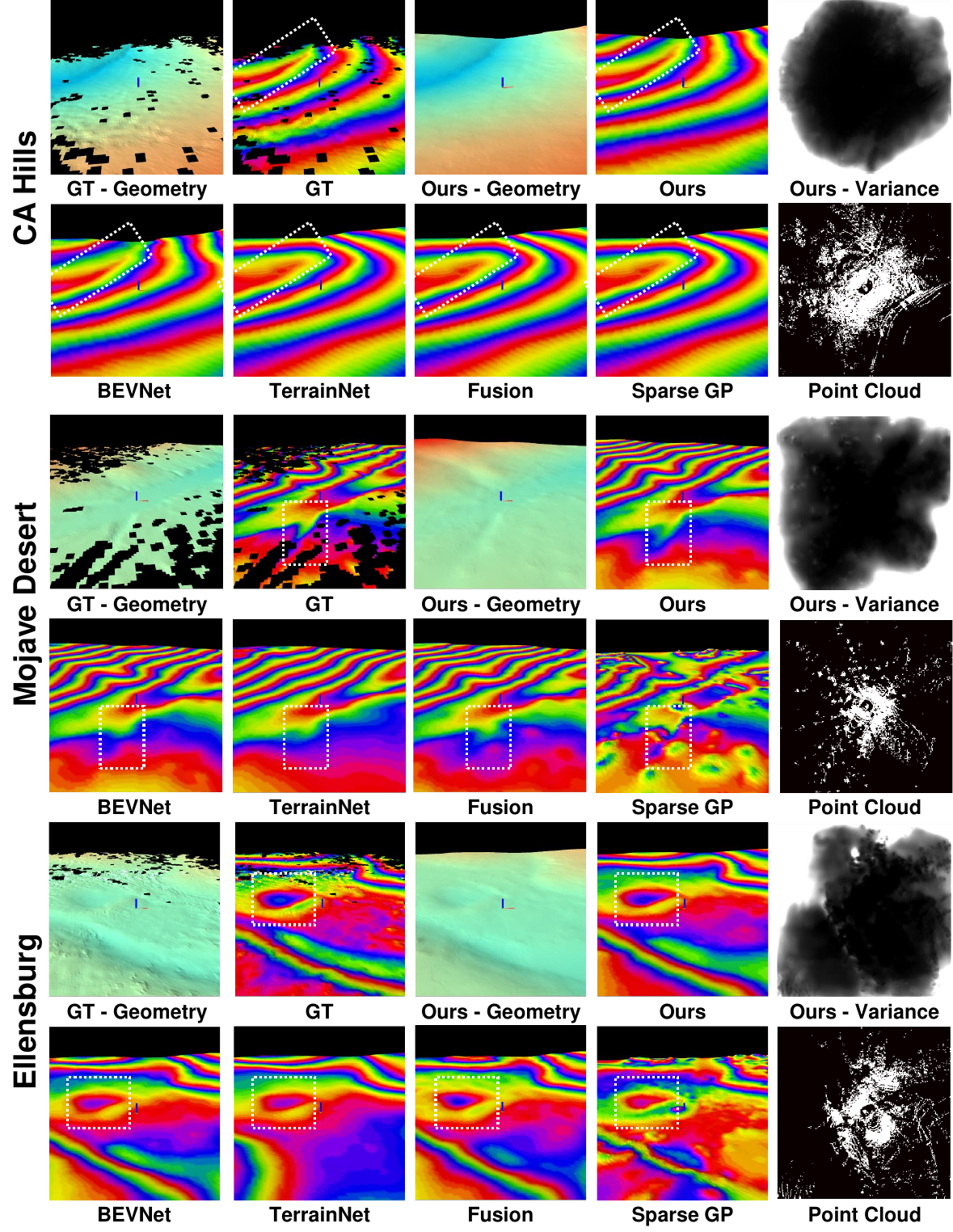}
  \vspace{-0.3cm}
  \caption{\textbf{Qualitative comparisons on the CA Hills, Mojave Desert, and Ellensburg sequences.} The predicted elevations are visualized in 3D and also color-coded with a repeated pattern.}
  \vspace{-0.4cm}
\label{suppfig:qualitative}
\end{figure}

\noindent\textbf{Comparisons with Baselines.}
Fig.~\ref{suppfig:qualitative} shows our predicted results and baseline methods on all three sequences.
As shown, baselines struggle with predicting negative obstacles in most cases, while our method successfully finds them by accurately estimating the slopes (white boxes on color-coded images).
Also, the predicted variances align with the point cloud observations.

\begin{figure}[t!]
  \centering
  \includegraphics[width=\textwidth]{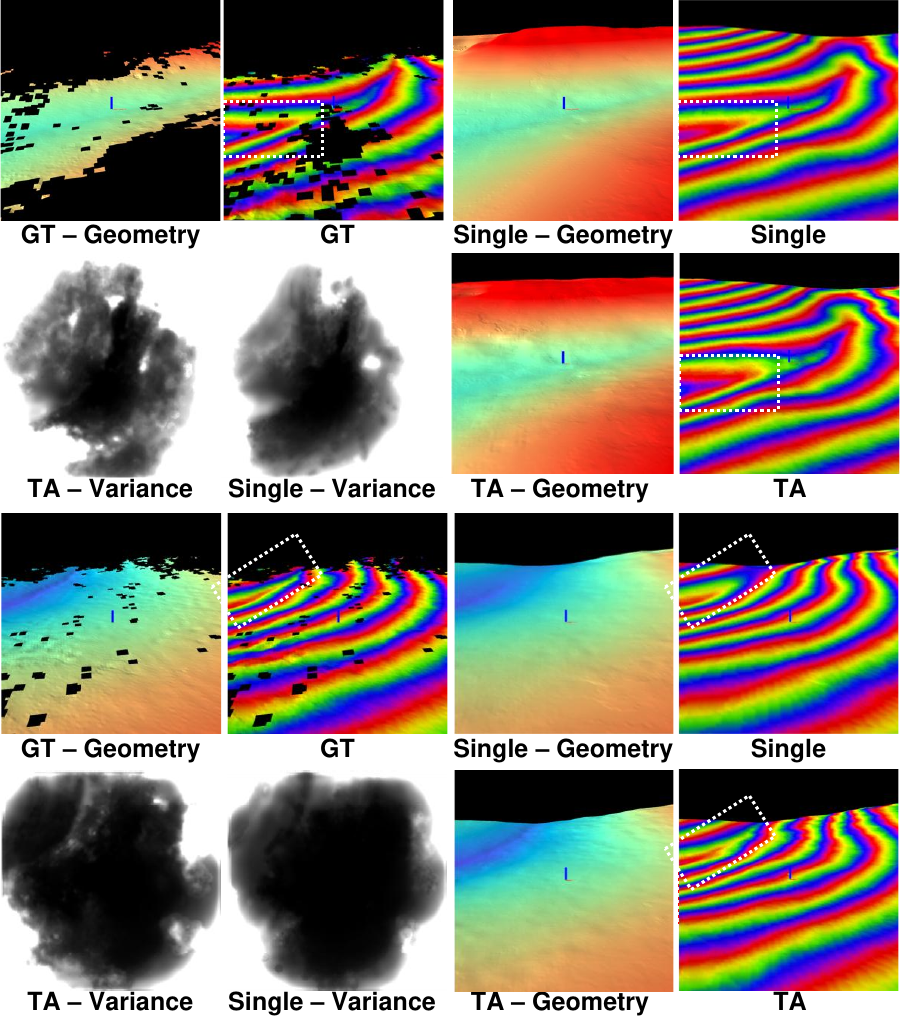}
  \vspace{-0.3cm}
  \caption{\textbf{Effectiveness of temporal aggregation in CA Hills.} The predicted elevations are visualized in 3D and also color-coded with a repeated pattern.}
  \vspace{-0.4cm}
\label{suppfig:qual_TA}
\end{figure}

\noindent\textbf{Effectiveness of Temporal Aggregation.}
Fig.~\ref{suppfig:qual_TA} demonstrates the effects of temporal aggregation by comparing the predictions with single-frame predictions.
As shown in geometry and color-coded visualizations (white boxes), the temporally-aggregated model preserves fine details from previous history and presents more accurate estimations.
Also, temporal aggregation lowers the uncertainty of previously observed regions, as shown in variance comparisons.

\begin{figure}[t!]
  \centering
  \includegraphics[width=\textwidth]{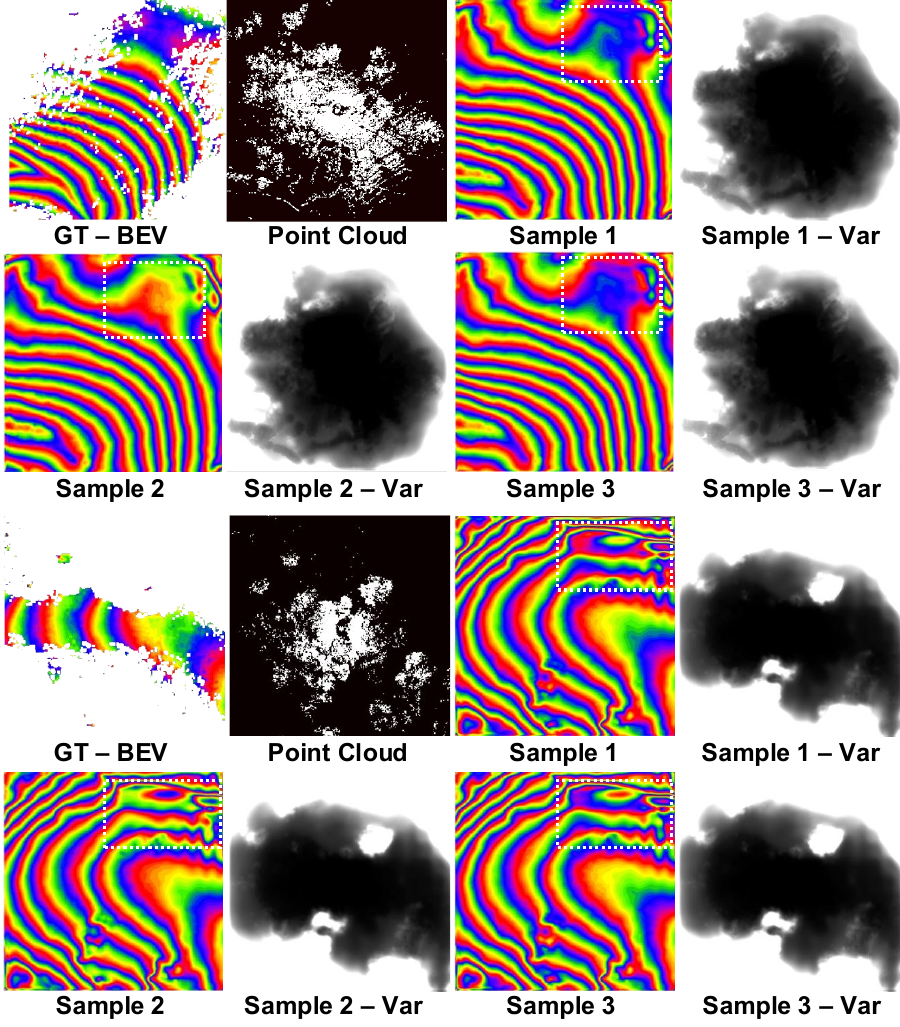}
  \vspace{-0.3cm}
  \caption{\textbf{Different elevation predictions across varying global latent $z$ samples from our method.}}
  \vspace{-0.4cm}
\label{suppfig:qual_samples}
\end{figure}

\noindent\textbf{Different Samples from Varying Global Latent.}
Our method is based on latent NPs where the decoder is conditioned on the global latent context vector $z$.
If we sample different $z$ from the predicted latent distribution, we can obtain different samples, especially in unobserved or highly uncertain areas.
From a planning and control perspective, this can be a significant benefit since the planner/controller can evaluate the trajectories on multiple possible scenarios.
Also, we can maintain spatial correlation between BEV cells because they are all generated from the same global latent $z$, rather than individually sampled in their own distributions.
As shown in Fig.~\ref{suppfig:qual_samples}, we can obtain various elevation predictions in uncertain or unobserved areas (i.e., white boxes) while preserving accurate elevation estimations in confident and observed regions.

\clearpage
\bibliography{reference}  

\end{document}